\documentclass{article}

\usepackage{PRIMEarxiv}

\usepackage[utf8]{inputenc} % allow utf-8 input
\usepackage[T1]{fontenc}    % use 8-bit T1 fonts
\usepackage{hyperref}       % hyperlinks
\hypersetup{
  pdftitle={Modeling Rapid Contextual Learning in the Visual Cortex with Fast-Weight Deep Autoencoder Networks},
  pdfauthor={Yue Li, Weifan Wang, Tai Sing Lee}
}

\usepackage{url}            % simple URL typesetting
\usepackage{booktabs}       % professional-quality tables
\usepackage{amsfonts}       % blackboard math symbols
\usepackage{nicefrac}       % compact symbols for 1/2, etc.
\usepackage{microtype}      % microtypography
\usepackage{lipsum}
\usepackage{graphicx}
\graphicspath{{media/}}     % organize your images and other figures under media/ folder
\usepackage{hyperref}

\usepackage{amsmath} 
\usepackage{subcaption}
\usepackage{natbib}
\usepackage{float}

\usepackage{fancyhdr}
\title{
Modeling Rapid Contextual Learning in the Visual\\
Cortex with Fast-Weight Deep Autoencoder Networks
}

\author{
{\Large Yue Li\textsuperscript{1}, 
Weifan Wang\textsuperscript{1}, 
Tai Sing Lee\textsuperscript{1}\textsuperscript{*}}\\[6pt]
{\large \textsuperscript{1}Carnegie Mellon University}\\[4pt]
\texttt{\{yueli4, weifanw, taislee\}@andrew.cmu.edu}
}

\begin{document}
\maketitle

\begin{abstract}
Recent neurophysiological studies have revealed that the early visual cortex can rapidly learn global image context, as evidenced by a sparsification of population responses and a reduction in mean activity when exposed to familiar versus novel image contexts. This phenomenon has been attributed primarily to local recurrent interactions, rather than changes in feedforward or feedback pathways—supported by both empirical findings and circuit-level modeling. Recurrent neural circuits capable of simulating these effects have been shown to reshape the geometry of neural manifolds, enhancing robustness and invariance to irrelevant variations. In this study, we employ a Vision Transformer (ViT)-based autoencoder to investigate, from a functional perspective, how familiarity training can induce sensitivity to global context in the early layers of a deep neural network. We hypothesize that rapid learning operates via fast weights, which encode transient or short-term memory traces, and we explore the use of Low-Rank Adaptation (LoRA) to implement such fast weights within each Transformer layer. Our results show that: (1) The proposed ViT-based autoencoder's self-attention circuit is performing a manifold transform similar to a neural circuit developed for modeling the familiarity effect. (2) Familiarity training induces alignment of latent representation in early layers with the top layer that contains global context information. (3) Familiarity training makes self-attention pay attention to a broader scope details in the remembered image context, rather than just the critical features for object recognition. (4) These effects are significantly amplified by the incorporation of LoRA-based fast weights. Together, these findings suggest that familiarity training can introduce global sensitivity to earlier layers in a hierarchical network, and that a hybrid fast-and-slow weight architecture may provide a viable computational model for studying the functional consequences of rapid global context learning in the brain. 

\noindent\textbf{Code:} \url{https://github.com/ron7li/familiarity_training}
\end{abstract}


\section{Introduction}

Visual perception operates in a world rich with ambiguity, where local features are interpreted in light of global context. The brain resolves this ambiguity by integrating information across spatial scales—leveraging long-range dependencies to disambiguate local signals. While global context is often modeled as top-down feedback—e.g., in analysis-by-synthesis or predictive coding frameworks —it can also reside within local circuits, dynamically modulated by global contextual signals \cite{rao1999predictive, lee2003hierarchical,  gilbert2013top, angelucci2017circuits, coen2015flexible}. This division of labor would reduce the informational burden on the feedback pathways, allowing the brain to mediate context with architectural efficiency.

Contextual modulation has long been recognized as a key feature of early visual processing, reflecting the influence of experience \cite{gilbert2012adult} and statistical priors derived from natural scenes \cite{geisler2008visual}. Yet recent neurophysiological findings suggest an even more striking phenomenon: global image context can be rapidly learned and expressed in early visual areas such as V1 and V2 \cite{yan2018bottom, huang2018neural}. Specifically, repeated exposure to a small set of natural images—typically 25 familiar images—over the first few days is sufficient to induce robust familiarity suppression in V2 \cite{huang2018neural}. Such rapid plasticity raises a fundamental question: how does the system acquire new contextual information so quickly, without suffering catastrophic overwriting of the existing knowledge or priors? 
One compelling explanation is that the visual system forms local and long-range subnetworks that can be selectively activated by familiar global contexts. When activated, these circuits suppress competing circuits, thus reducing overall activity while sharpening population activities. This architecture would allow multiple contexts to be encoded in parallel, enhancing efficiency and specificity without requiring wholesale modification of feedforward or feedback weights.

Inspired by these findings, we investigate how similar forms of rapid contextual learning might be instilled in artificial neural networks, particularly within early layers, without disrupting previously acquired knowledge. We ask: what kind of architectural and learning mechanisms are required to enable such rapid, context-sensitive plasticity? And what are the functional consequences of this ability—for robustness, invariance, and specificity in visual representation?

To explore these questions, we develop a biologically motivated analysis-by-synthesis framework using autoencoder-based networks trained under self-supervised protocols that mimic visual familiarity exposure. Within this architecture, we introduce a mechanism for temporary memory in the form of fast weights—transient synaptic changes that allow rapid adaptation without compromising the stability of slower, structural memory circuits. We implement these fast weights using low-rank adaptation (LoRA) \cite{hu2022lora}, a method originally developed for parameter-efficient fine-tuning, and repurpose and modify it here as a computational analog of rapid, reversible plasticity. By embedding fast weights into the manifold-transform components of the network, we enable it to encode global contextual signals with minimal interference to core statistical priors. 

We test the hypothesis that this form of familiarity training can induce global context sensitivity in early-layer representations. Our results show that these early representations become more aligned with the representation of the highest and most global layer, suggesting a functional reorganization of the representational geometry that reflects global sensitivity in the early layers. Furthermore, We found that networks augmented with LoRA exhibit faster and more robust alignment between early-layer representations and higher-level global representations. Moreover, their attention mechanisms display more coherent and consistent figure–ground segregation, reflecting improved contextual awareness. Finally, we demonstrate that the learned manifold transformations resemble those implemented by recurrent circuit models of familiarity in early visual cortex, compressing irrelevant variation while preserving object-discriminative dimensions of the neural manifold.

Together, these findings point toward a principled mechanism by which global contextual memory can be rapidly and efficiently encoded in early visual representations, with implications for both  understanding the brain and for the design of adaptive neural architectures.

% {\em This paper explores the hypothesis that familiarity effects observed in the mammalian visual cortex reflect a form of temporary memory. We investigate how to implement this mechanism in deep learning models, especially the Vision Transformer models (ViT) \cite{dosovitskiy2020image}, by using LoRA \cite{hu2022lora} as a form of temporary memory, enabling more efficient visual processing. Besides, we explore whether the intermediate layers of the model can achieve similar performance and characteristics to the final layer through familiarity training.} 

\section{Background and Related Works}

\subsection{Neuroscience of familiarity learning}
Familiarity learning refers to the phenomenon—well documented in inferotemporal cortex (ITC) \cite{meyer2014image, fahy1993neuronal, xiang1998differential, mruczek2007context, sobotka1993investigation, freedman2006experience, woloszyn2012effects} and, more recently, in early visual cortex \cite{huang2018neural}—where repeated exposure to a set of stimuli produces a suppression of average population responses, a sharpening of single-neuron tuning curves, and a sparsification of population activity relative to responses to novel stimuli \cite{woloszyn2012effects, freedman2006experience, tang2018large} (\ref{fig:framework}B and \ref{fig:framework}C). 
\cite{huang2018neural} further showed that neurons with localized receptive fields become sensitive to the global context of familiar images. 
The timing of these effects suggests that they are mediated primarily by recurrent circuitry within V2 rather than by feedback from higher visual areas. Together, these observations point to a rapid plasticity mechanism in early visual cortex that modifies local recurrent circuits in each visual area along the visual hierarchy to encode relatively global, at a scale appropriate at each level,  context.

\subsection{Neural circuit modeling and manifold transform}
\cite{wang2025manifold} developed a V1-based neural circuit model, using Hebbian learning and standard V1 circuit motifs, that accounts for these familiarity effects (\ref{fig:framework}A,B,C). The central idea is that repeated exposure to a particular global image context drives the formation of local excitatory subnetworks that encode that context via Hebbian plasticity. Such subnetworks support pattern completion when inputs are occluded or corrupted by noise. By analyzing the recurrent circuit’s input–output neural activities, \cite{wang2025manifold} showed that the circuit implements a manifold transform: it compresses task-irrelevant (within-context) variability while preserving distinctions between different global contexts.

While \cite{wang2025manifold} revealed this manifold-transform view in a biologically grounded, single-area model, several questions remain. Can such circuits be generalized to hierarchical networks? How does sensitivity to global context evolve across layers, and where do similarities or divergences from biological circuits emerge? Are these phenomena specific to Hebbian-trained, biologically realistic circuits, or do they generalize to modern deep networks trained with backpropagation trained under similar conditions?

To address these questions, we construct a hierarchical autoencoder that uses the image encoder from CLIP \cite{radford2021learning} as a backbone—an architecture shown to align well with the hierarchical organization of human visual cortex. Functionally, the encoder–decoder pair mirrors (to a degree) feedforward and feedback pathways in the brain. This setup affords two advantages: (i) it leverages a powerful state-of-the-art encoder that can be trained or fine-tuned with backpropagation, and (ii) it allows us to introduce LoRA as a form of fast memory. If a deep network trained in this way exhibits analogous familiarity effects and manifold transforms layer-by-layer, this would support the generality of the underlying principles beyond biological circuits.

\subsection{LoRA and fast weights}
%Rapid plasticity effect has been observed in the early visual cortex. Measurable changes in short-term sensitivity to global context have been observed to develop within a recording session (stimulus-specific repetition suppression \cite{Peter2021, olson2022}, consolidated into memory over three days of training \cite{huang2018neural}. 
Familiarity training in cortex appears to be mediated by rapid plasticity: 
within a few days of repeated exposure, V2 neurons exhibit familiarity suppression that depends on global context \cite{huang2018neural}.
Rapidly learning new episodic information, however, risks catastrophic forgetting of existing knowledge. Two influential, complementary solutions have been proposed: the complementary learning systems framework \cite{mcclelland1995there} and fast vs. slow weights \cite{hinton1987using}. Here we will focus our study on the latter.  

Fast and slow weights separate learning across timescales. Slow weights encode stable, semantic knowledge (synaptic consolidation), while fast weights support rapid, context-dependent, episodic-like adaptation without overwriting long-term memory. \cite{ba2016using} formalized fast weights in RNNs by introducing a secondary, rapidly updated, auto-associative weight matrix—functionally reminiscent of a Hopfield network layered atop the RNN \cite{hopfield1982neural}.

We propose an alternative implementation of fast weights using LoRA (Low-Rank Adaptation). LoRA is a parameter-efficient fine-tuning method in which low-rank, additive adapters are trained while the base model remains frozen. This modularity allows different LoRA modules to be swapped for different tasks or contexts without retraining the entire model. Because LoRA adapters are low rank and explicitly decoupled from the slow (frozen) parameters, they confer both data and training efficiency when learning new contexts—precisely the desiderata of fast weights.

LoRA is commonly applied to the self-attention circuitry of transformer-based networks \cite{vaswani2017attention}. Since transformers capture both local and long-range dependencies—and attention has been argued to be mathematically related to modern Hopfield networks in certain regimes \cite{ramsauer2020hopfield}—LoRA-equipped transformers offer a scalable substrate for fast-weight-like mechanisms. This motivates our use of LoRA within a ViT-based \cite{dosovitskiy2020image} autoencoder to investigate the functional consequences of familiarity training in a hierarchical visual system. Unlike typical LoRA applications that freeze the entire pretrained model, we adopt a partial-freezing strategy: only the modules into which LoRA is inserted are frozen, while surrounding layers remain trainable. This design aims to better emulate local circuit-level plasticity in the brain and reflects the coexistence of slow, stable pathways and fast, adaptive components observed in cortical microcircuits.

\section{Approach}

This section outlines our approach to investigating the impact of familiarity training on neural representation, particularly in early visual areas of a hierarchical visual system or in the early layers of a neural network. We aim to evaluate two key hypotheses: (1) familiarity training introduces global contextual information into early layers, and (2) it alters the manifold transformation at each level to compress irrelevant variant dimensions while preserving distinctions between different image contexts.

\begin{figure}[t]
    \centering
    \includegraphics[width=1\linewidth]{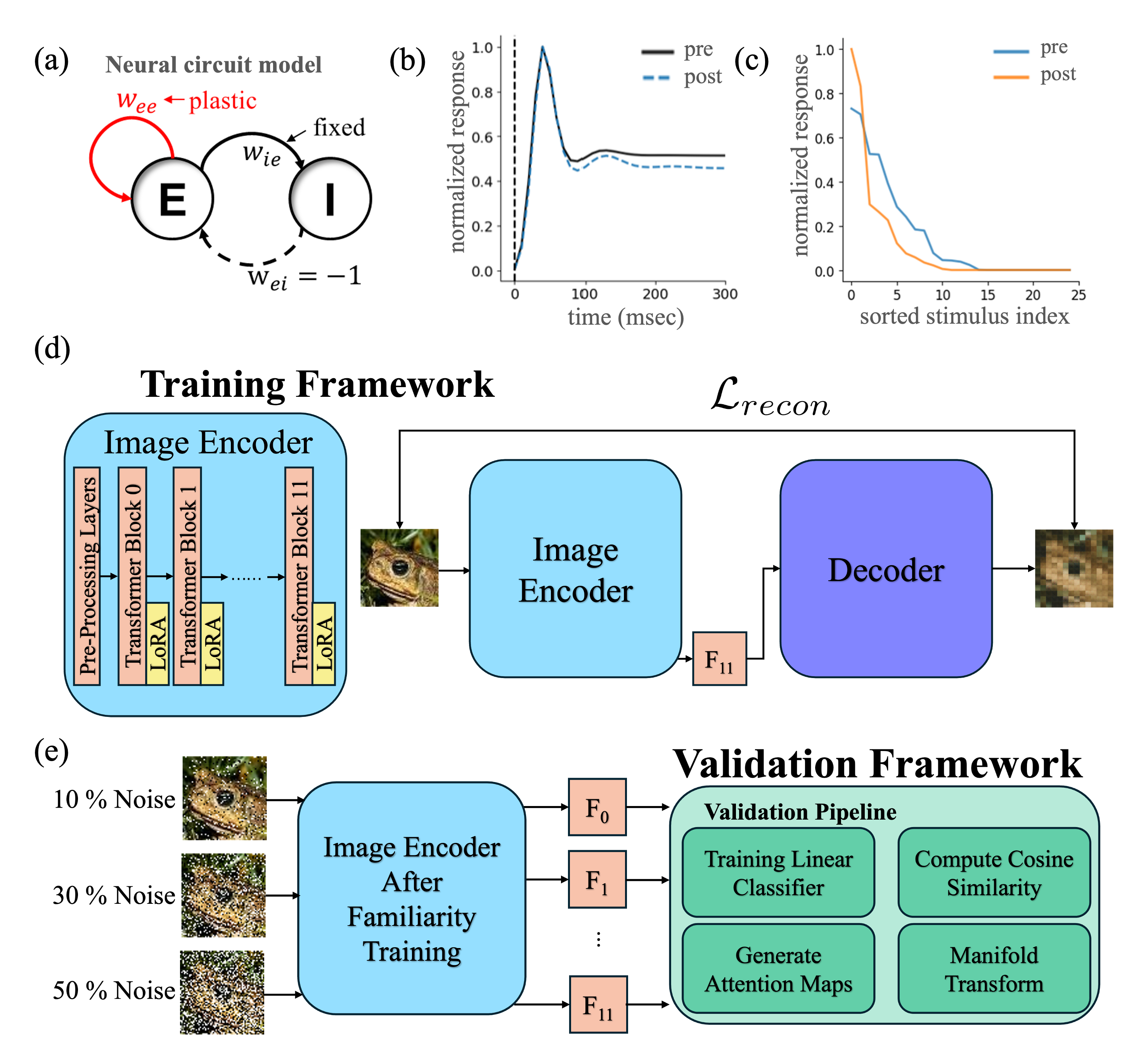}
    \captionsetup{width=1\linewidth}
    \caption{\textbf{(a)} Neural Recurrent Circuit Model for Familiarity training (see Wang et al (2025)) 
            \textbf{(b)} Familiarity Suppression -- population averaged responses to familiar images were found to be suppressed relative to their averaged responses to novel images. (see Huang et al. (2018))  
            \textbf{(c)} Sharpening of the population response to familiar images relative to novel images suggests the neural representation of familiar images becomes sparsified.  
            \textbf{(d)} ViT-based Autoencoder Framework for modeling familiar training using deep networks. 
            \textbf{(e)} Testing and Evaluation Framework: The network was trained with clean images but tested with noisy images with different levels of noise. Four types of evaluations were performed to compare with the recurrent circuit, to assess the impact of familiarity training on manifold transform, development of global context sensitivity in early layers of the Image Encoder, an analog of the early visual cortex, and the modification of attention focus in the self-attention map. }
    \label{fig:framework}
\end{figure}

We simulate familiarity training via passive exposure using an autoencoder with a reconstruction loss (\ref{fig:framework}D). The encoder is a pretrained CLIP ViT-B/16 model \cite{radford2021learning}, selected for its alignment with the hierarchical organization of the primate visual cortex \cite{yang2024brain}. The decoder is a lightweight transformer with 8 layers, 8 attention heads, and a 512-dimensional embedding. It reconstructs the input image from the encoder’s highest-level latent representation.

The reconstruction loss combines mean squared error (MSE) and L1 loss:

\begin{equation}
\label{eq:loss}
\mathcal{L}_{\text{recon}} = \mathcal{L}_{\text{MSE}} + \lambda \mathcal{L}_{\text{L1}}
\end{equation}

where $\lambda = 0.1$ controls the L1 term. Gradients from this loss are backpropagated through both encoder and decoder, enabling end-to-end training.

The model is trained for 100 epochs with a batch size of 4, a learning rate of $1\times10^{-4}$, and the AdamW optimizer. We assume that self-supervised learning under this reconstruction objective mimics familiarity learning under passive exposure, consistent with predictive coding and analysis-by-synthesis frameworks.

Since self-attention enables long-range dependencies within layers, we hypothesize that it can serve as a functional analog of biological recurrent circuits mediating contextual interactions. To test this, we compare the input-output mappings of each transformer block with those of recurrent circuits as described in \cite{wang2025manifold}.

We performed four analytical evaluation on the networks' activities (\ref{fig:framework}E) to assess:
1. Whether familiarity training is performing a manifold transform similar to the neural recurrent circuit.
2. Whether incorporating LoRA enhances the effects of familiarity on manifold geometry and context sensitivity.
3. Whether familiarity training induces global context sensitivity in the different self-attention layers in the ImageEncoder, particularly the early layers.
4. In what way does familiarity trainig change the self-attention computation in the network?  
 
Following \cite{wang2025manifold}, we will test the network with familiarity training of 4 global image contexts randomly sampled from distinct categories in the Tiny ImageNet-200 dataset \cite{le2015tiny}. Each image context can be corrupted by salt-and-pepper noise at 10\%, 30\%, or 50\% levels, yielding 12 image-noise context combinations. 
The network is trained only on clean images but tested on all image-noise combination contexts, with 10 samples drawn from each context. In each validation run, we evaluate the network on 10 samples per context, using only one specific noise level across all image contexts.

We will compare two network architectures:
1. Without LoRA: the encoder is fine-tuned via full-parameter optimization.
2. With LoRA: only the original $\mathbf{W}_Q$, $\mathbf{W}_K$, and $\mathbf{W}_V$ in the self-attention blocks are frozen, while the LoRA modules and all other model parameters remain trainable.

% In the LoRA-equipped network, low-rank ($r=8$) adapters are inserted into the projection matrices $\mathbf{W}_Q$, $\mathbf{W}_K$, and $\mathbf{W}_V$ to serve as “fast weights.” These modules selectively adapt the attention mechanism during familiarity training, while other network parameters remain learnable.
In the LoRA-equipped network, the frozen projection matrices serve as slow weights that preserve long-term visual priors. Into these matrices, we insert low-rank ($r=8$) LoRA adapters that act as fast synaptic adjustments, selectively modulating the attention mechanism during familiarity training. All other model parameters—including those outside the attention projections—remain trainable and are functionally integrated into a distributed fast-weight subsystem, supporting task-specific adaptation beyond the LoRA injection points.

To analyze layer-wise behavior, we extract the \texttt{[CLS]} token embedding from each layer as a summary of that layer’s representational activity.

\section{Results}

\subsection{Impact of Familiarity Training on Manifold Transformations}

To evaluate how familiarity training alters representational geometry, we compare the manifold transformations induced by the recurrent circuit model and the transformer-based autoencoder. Following the methodology of Wang et al. (2025), we compute relative distances to characterize intra-context variation (within an image across noise levels) and inter-context separability (across different images). 

\begin{figure}[htbp]
    \centering
    \includegraphics[width=1\linewidth]{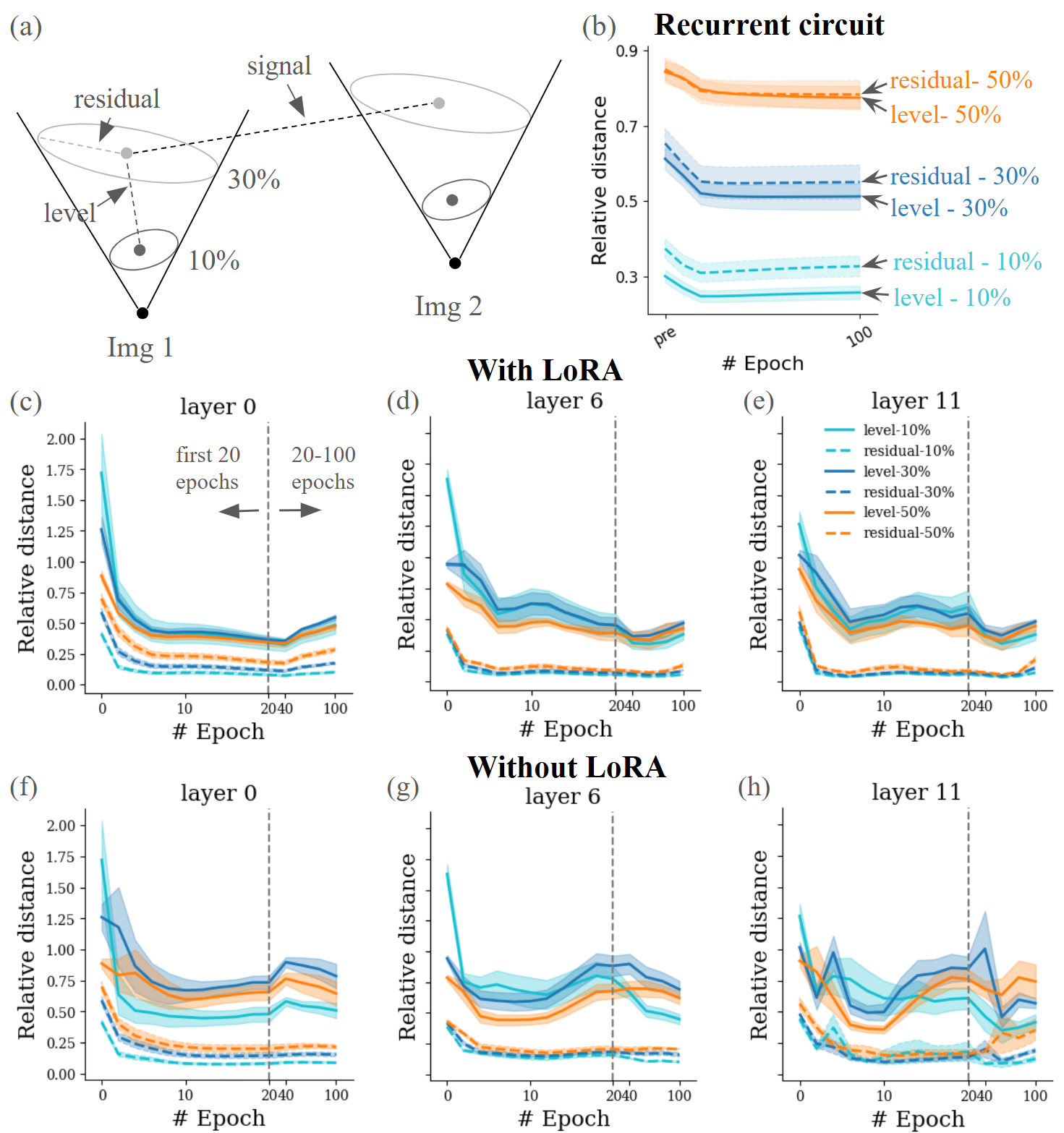}
    \captionsetup{width=1\linewidth}
    \caption{\textbf{(a)} Illustration of the manifold geometry for two image contexts and the associated noise-cone. The cone’s cross-section ellipses show sample variability at each noise level, and its axis marks increasing noise. Distances across noise levels and within each level quantify how the manifold expands or contracts. \textbf{(b)} Learning curve of relative distances in the recurrent neural circuit model. \textbf{(c-e)} ViT encoder with LoRA. \textbf{(f-h)} ViT encoder without LoRA. Training reduces both relative level and residual distances, indicating manifold compression.}
    \label{fig:relative_distance}
\end{figure}

Figure~\ref{fig:relative_distance}A illustrates this analysis. Each cone represents the manifold formed by samples of a given image across multiple noise levels. Ellipsoids within the cone correspond to distributions of representations at individual noise levels. 

We define two key relative distance metrics:

- \textit{Relative Level Distance} measures the distance between representations of the same image at adjacent noise levels, normalized by the average inter-image distance at the same noise level.
- \textit{Relative Residual Distance} measures the spread (variance) of samples within a single image-noise cluster, also normalized by the inter-image distance.

Let $\mathbf{z}_{l,n,k}$ denote the representation of sample $k$ from image $l$ at noise level $n$. These metrics are formalized as:

\begin{equation}
R_{\text{level}}(k, n, l) = \sqrt{\frac{\sum_{k'}\| \mathbf{z}_{k,n,l} - \mathbf{z}_{k',n-1,l} \|^2}{\sum_{m\neq l,j} \| \mathbf{z}_{k,n,l} - \mathbf{z}_{j,n,m} \|^2}},
\label{eq:level_dist}
\end{equation}

\begin{equation}
R_{\text{residual}}(k, n, l) = \sqrt{\frac{\sum_{i\neq k}\| \mathbf{z}_{k,n,l} - \mathbf{z}_{i,n,l} \|^2}{\sum_{m\neq l,j} \| \mathbf{z}_{k,n,l} - \mathbf{z}_{j,n,m} \|^2}}.
\label{eq:residual_dist}
\end{equation}

In the recurrent neural circuit model, familiarity training leads to a reduction in both relative level and residual distances, indicating compression of the variant manifold and improved representational efficiency (Figure~\ref{fig:relative_distance}B). 

Figures~\ref{fig:relative_distance}C–E and F–H show that similar effects emerge in the transformer-based autoencoder. Across training epochs, we observe consistent decreases in relative distances, particularly in early layers of the encoder. The compression is more pronounced when LoRA is incorporated, suggesting that LoRA-enhanced fast weights amplify the effects of familiarity training.

Overall, these findings indicate that familiarity training compresses task-irrelevant variation (e.g., noise-induced variability) while preserving task-relevant distinctions (e.g., image identity). This pattern mirrors the representational geometry observed in trained neural circuits in early visual areas.

%\subsection{Enhanced Context Discriminability via LoRA}

A consequence of variant manifold compression is improved discriminability between different image-noise contexts. To assess this, we trained a linear decoder to classify joint image-noise contexts based on representations from each encoder layer.

\begin{figure}[htbp]
    \centering

    \begin{subfigure}[t]{0.48\linewidth}
        \centering
        \includegraphics[width=\linewidth]{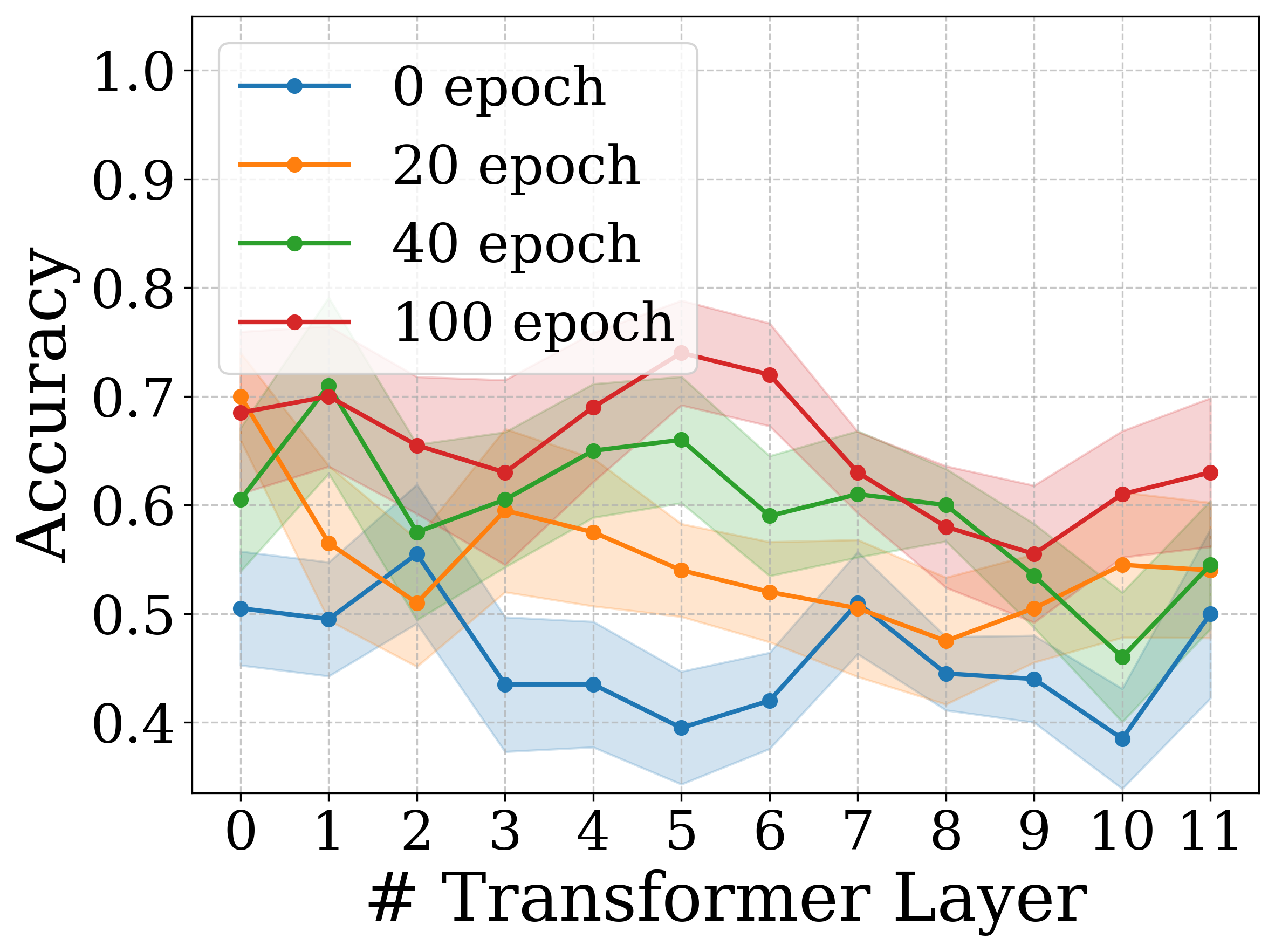}
        \caption{Without LoRA}
        \label{fig:acc_wo_lora}
    \end{subfigure}
    \hfill
    \begin{subfigure}[t]{0.48\linewidth}
        \centering
        \includegraphics[width=\linewidth]{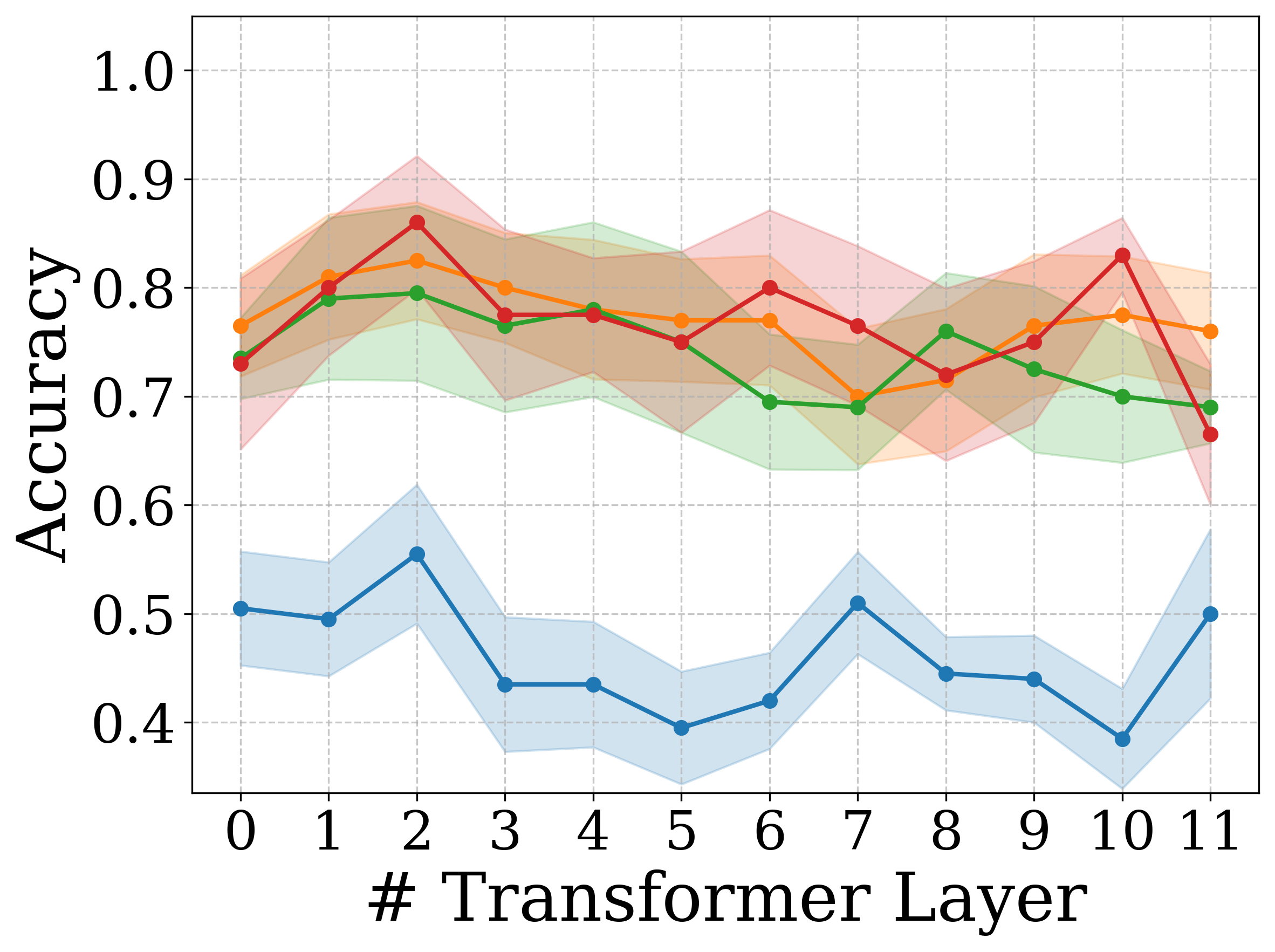}
        \caption{With LoRA}
        \label{fig:acc_w_lora}
    \end{subfigure}

    \caption{Evolution of the classification accuracy of the linear decoder applied to every transformer layer output during familiar training. The test stimuli were the different image contexts remembered corrupted with 30 \% salt and pepper noises. Decoding accuracy reflects the average of the discriminability or untangling of any two trained image contexts. Networks with LoRA fast weights ({\bf b}) exhibit stronger improvement of discriminability than networks without LoRA ({\bf a}). }
    \label{fig:acc_wo_w_lora}
\end{figure}

Figure~\ref{fig:acc_wo_w_lora} compares discriminability for the LoRA and non-LoRA conditions. The LoRA-enhanced model achieves consistently higher classification accuracy, supporting the hypothesis that fast-weight mechanisms improve representational separation of the different familiar image contexts.

Interestingly, we find that average discriminability is relatively stable across layers, suggesting that while familiarity training compresses variance within noise cones, the separation between context manifolds is preserved throughout the hierarchy.

\subsection{Impact of Familiarity Training on Global Context Sensitivity}

We next examine whether familiarity training induces global context sensitivity in the early layers of the autoencoder network. We operationalize global context as the representation in the top layer (layer 11) of the image encoder, which we assume encodes the most abstract and holistic features of the input. For each sample from an image-noise context, we compute the cosine similarity between its layer-wise latent representations and the corresponding clean image’s top-layer representation, hereafter referred to as the “global context representation.”

\begin{figure}[htbp]
  \centering
  \begin{subfigure}[b]{0.48\columnwidth}
    \centering
    \includegraphics[width=\linewidth]{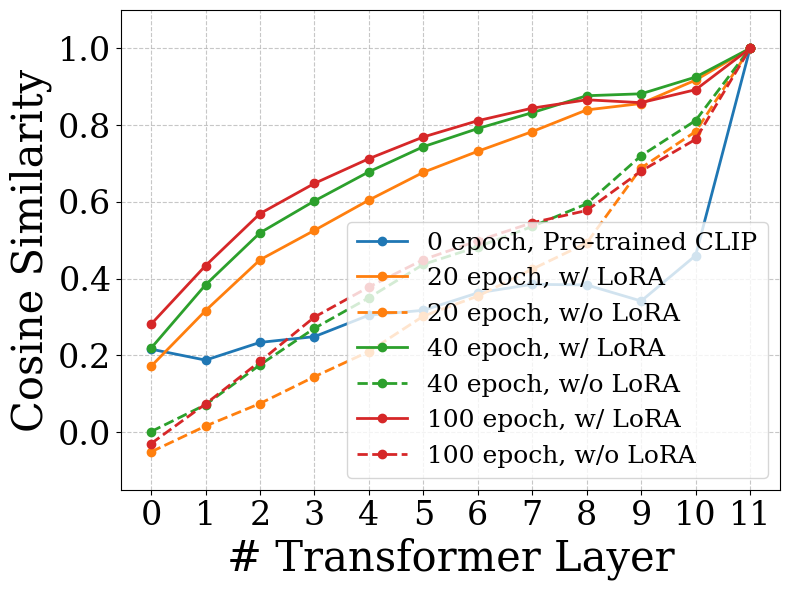}
    \caption{Noise Ratio = 0.0 (Clean)}
    \label{fig:cos_clean}
  \end{subfigure}\hfill
  \begin{subfigure}[b]{0.48\columnwidth}
    \centering
    \includegraphics[width=\linewidth]{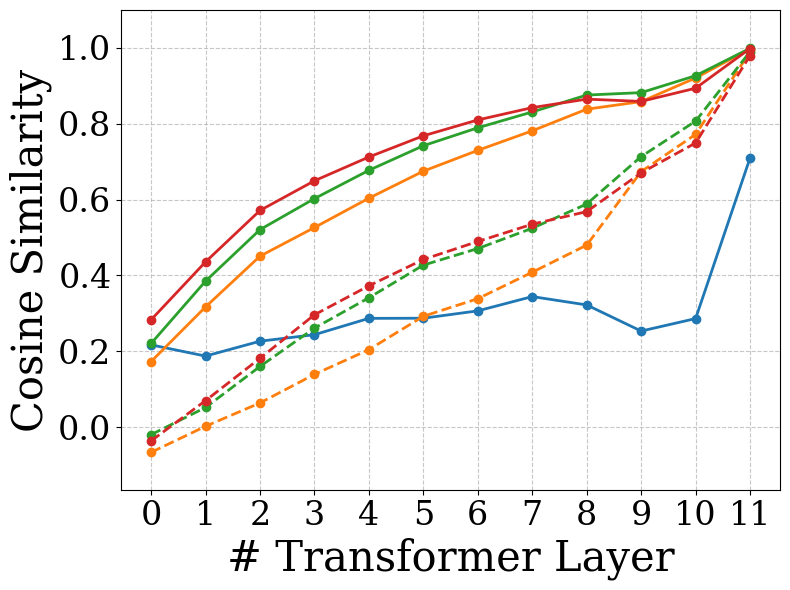}
    \caption{Noise Ratio = 0.1}
    \label{fig:cos_01}
  \end{subfigure}

  \vspace{2mm}

  \begin{subfigure}[b]{0.48\columnwidth}
    \centering
    \includegraphics[width=\linewidth]{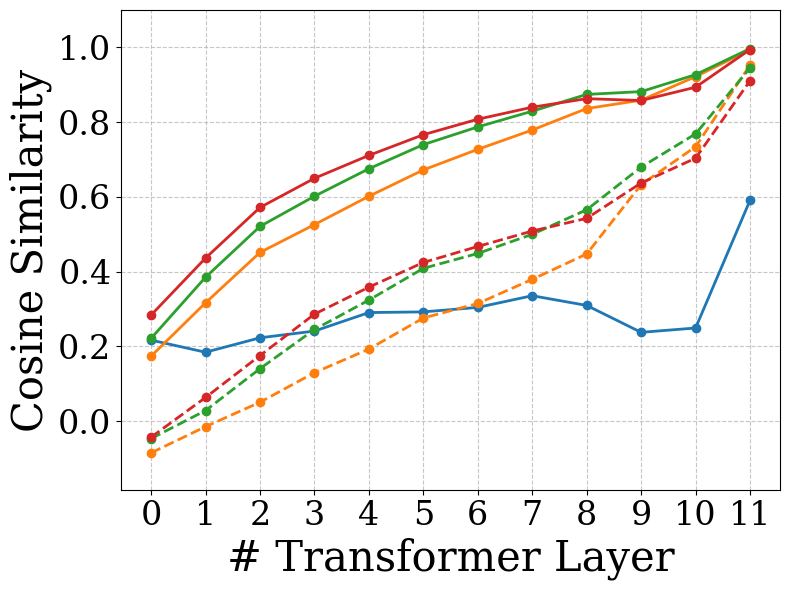}
    \caption{Noise Ratio = 0.3}
    \label{fig:cos_03}
  \end{subfigure}\hfill
  \begin{subfigure}[b]{0.48\columnwidth}
    \centering
    \includegraphics[width=\linewidth]{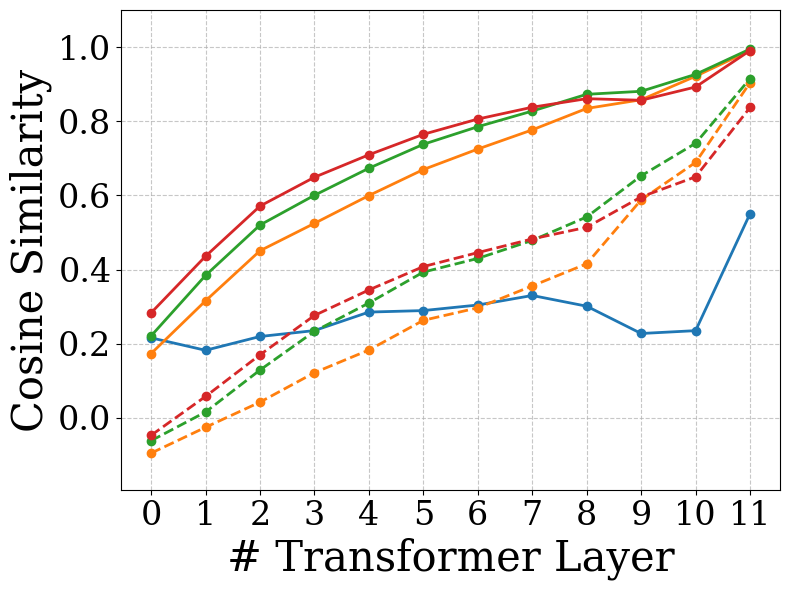}
    \caption{Noise Ratio = 0.5}
    \label{fig:cos_05}
  \end{subfigure}

  \caption{Cosine similarity between the class token embedding of every layer and the top (i.e. the 11th) layer of the Image Encoder which reflects global context.  LoRA network's layers consistently experience a progressive increase in the alignment with the global context, starting in layer 1. This alignment is considerably weaker for networks without LoRA, particularly in the early layers. The increase in alignment with the global context seems to be stabilized by 20 epochs and is essentially the same regardless of noise level. }
  \label{fig:feature_cos_all}
\end{figure}

Figure~\ref{fig:feature_cos_all} shows how these similarities evolve over the course of familiarity training. As training progresses, we observe an increasing alignment between early-layer representations and the global context representation. This indicates that the early layers of the network progressively acquire global contextual sensitivity.

This effect is consistently stronger and emerges more rapidly in the LoRA-enhanced model compared to the non-LoRA baseline, across all noise levels. Notably, the alignment is robust to noise perturbations up to 50\%, suggesting that LoRA facilitates the formation of globally aligned, noise-invariant features throughout the network hierarchy.

These findings support the hypothesis that familiarity training enhances the global context sensitivity of early layers, and that LoRA accelerates and amplifies this effect via a fast-weight plasticity mechanism.

Interestingly, this increase in global alignment closely tracks the compression of variant manifolds described earlier. By epoch 20, manifold compression (as reflected by reduced level and residual distances) has largely stabilized, coinciding with the plateauing of global alignment, particularly in higher layers. Additionally, the alignment is consistently stronger in the LoRA condition, paralleling the stronger manifold compression observed with LoRA.

These correlations suggest a potential mechanistic link between variant manifold compression and the emergence of global context sensitivity. However, further experimentation is needed to establish causality and directionality between these phenomena.

\subsection{Impact of Familiarity Training on Self-Attention}

To better understand why LoRA enhances familiarity-driven manifold compression, untangling, and global alignment, we investigate the effect of familiarity training on the network's self-attention mechanisms. Attention score maps derived from the \texttt{[CLS]} token embedding serve as a window into how information is aggregated into global representations and which image patches contribute most prominently to this process.

\begin{figure*}[htbp]
    \centering
    \begin{subfigure}[t]{0.48\textwidth}
        \centering
        \includegraphics[width=1\linewidth]{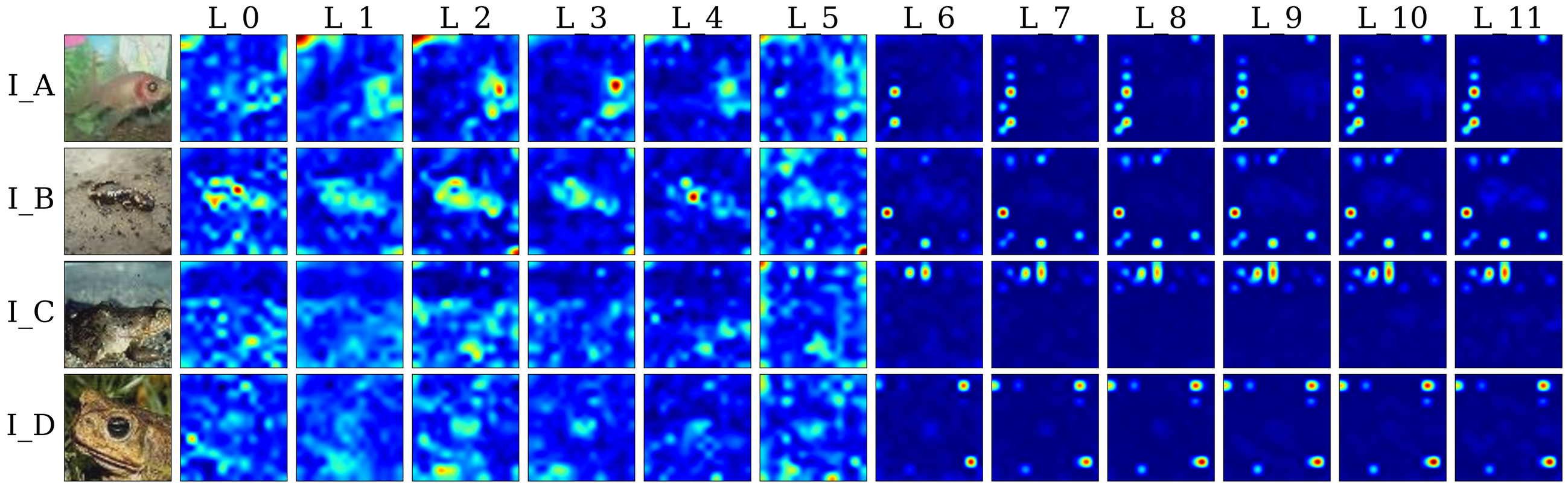}
        \caption{Pretrained model, clean input}
        \label{fig:map_before_train_clean}
    \end{subfigure}
    \hfill
    \begin{subfigure}[t]{0.48\textwidth}
        \centering
        \includegraphics[width=1\linewidth]{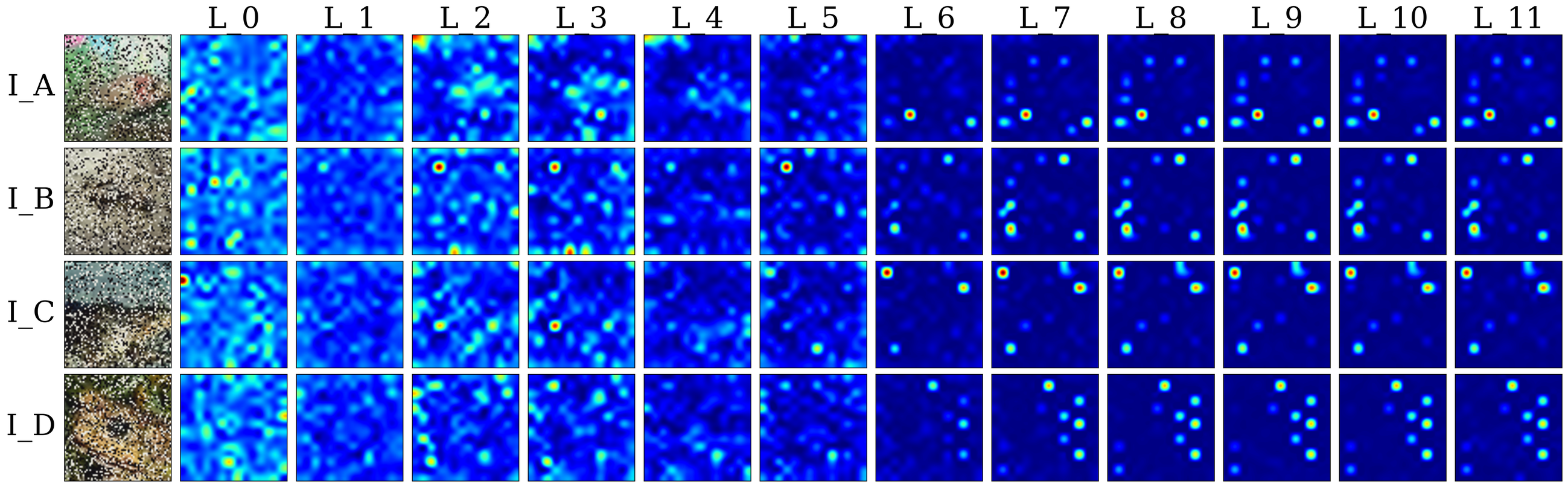}
        \caption{Pretrained model, noisy input (30\% noise)}
        \label{fig:map_before_train_noise}
    \end{subfigure}

    \vspace{2mm}

    \begin{subfigure}[t]{0.48\textwidth}
        \centering
        \includegraphics[width=1\linewidth]{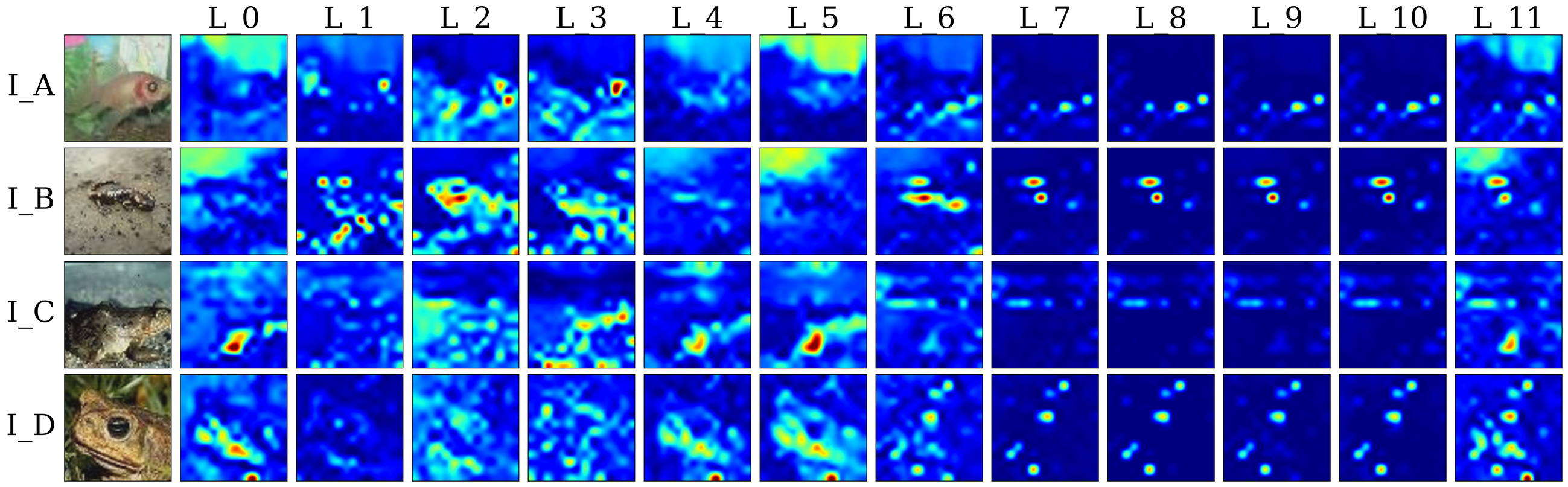}
        \caption{After training without LoRA, clean input}
        \label{fig:wo_lora_map_after_train_clean}
    \end{subfigure}
    \hfill
    \begin{subfigure}[t]{0.48\textwidth}
        \centering
        \includegraphics[width=1\linewidth]{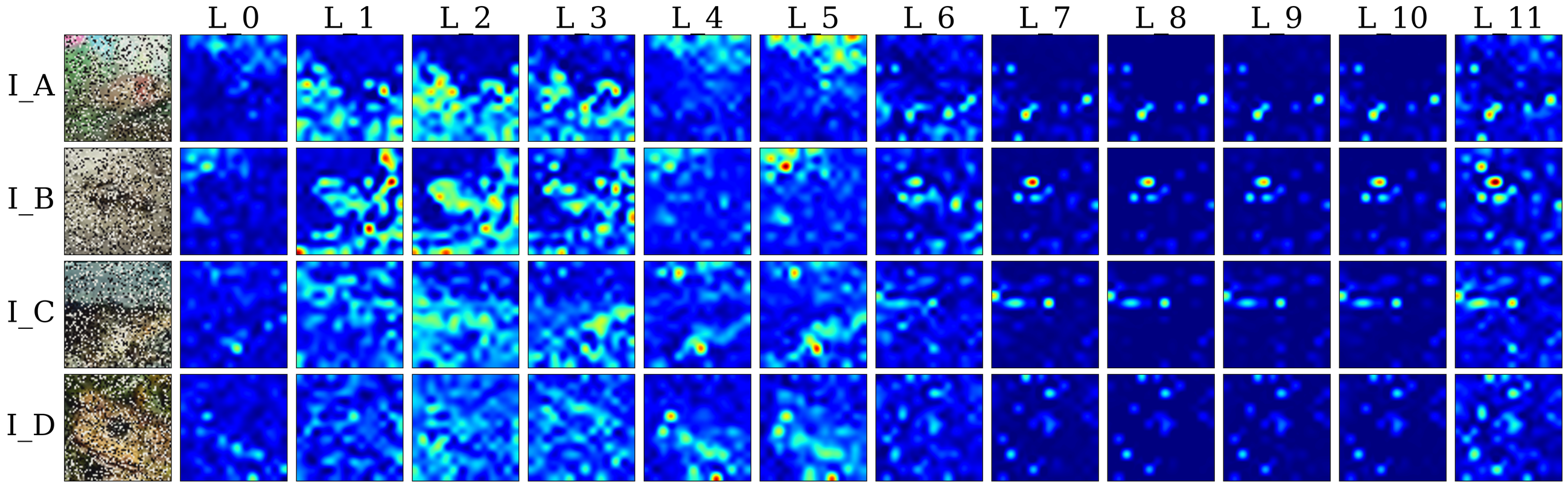}
        \caption{After training without LoRA, noisy input (30\%)}
        \label{fig:wo_lora_map_after_train_noise}
    \end{subfigure}

    \vspace{2mm}

    \begin{subfigure}[t]{0.48\textwidth}
        \centering
        \includegraphics[width=1\linewidth]{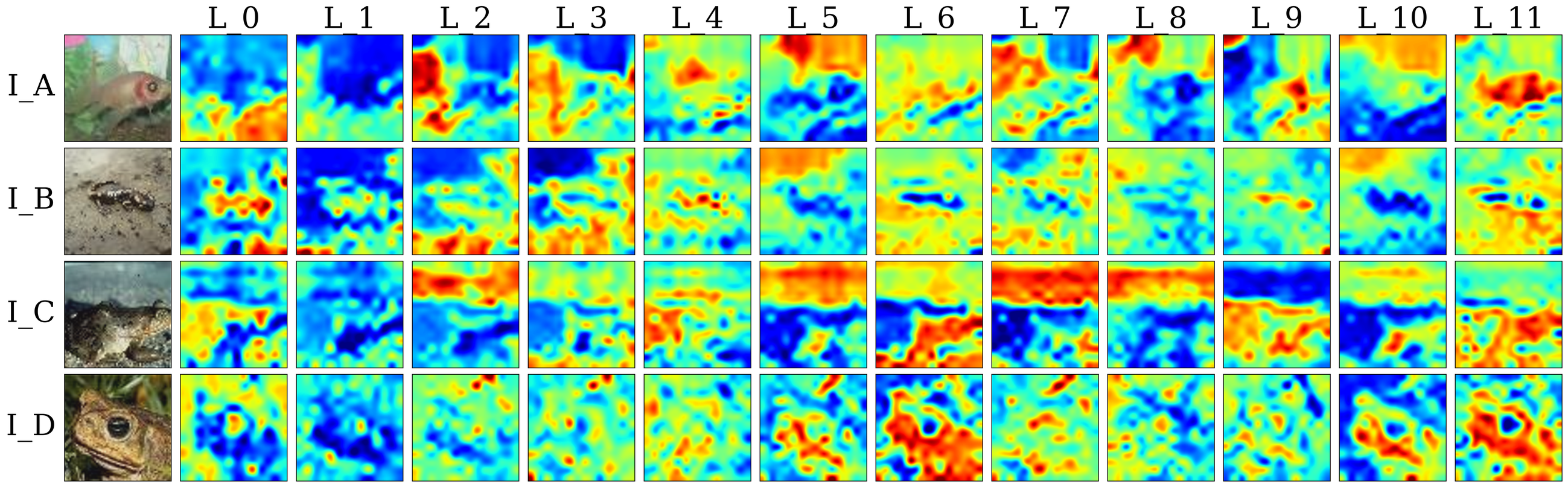}
        \caption{After training with LoRA, clean input}
        \label{fig:w_lora_map_after_train_clean}
    \end{subfigure}
    \hfill
    \begin{subfigure}[t]{0.48\textwidth}
        \centering
        \includegraphics[width=1\linewidth]{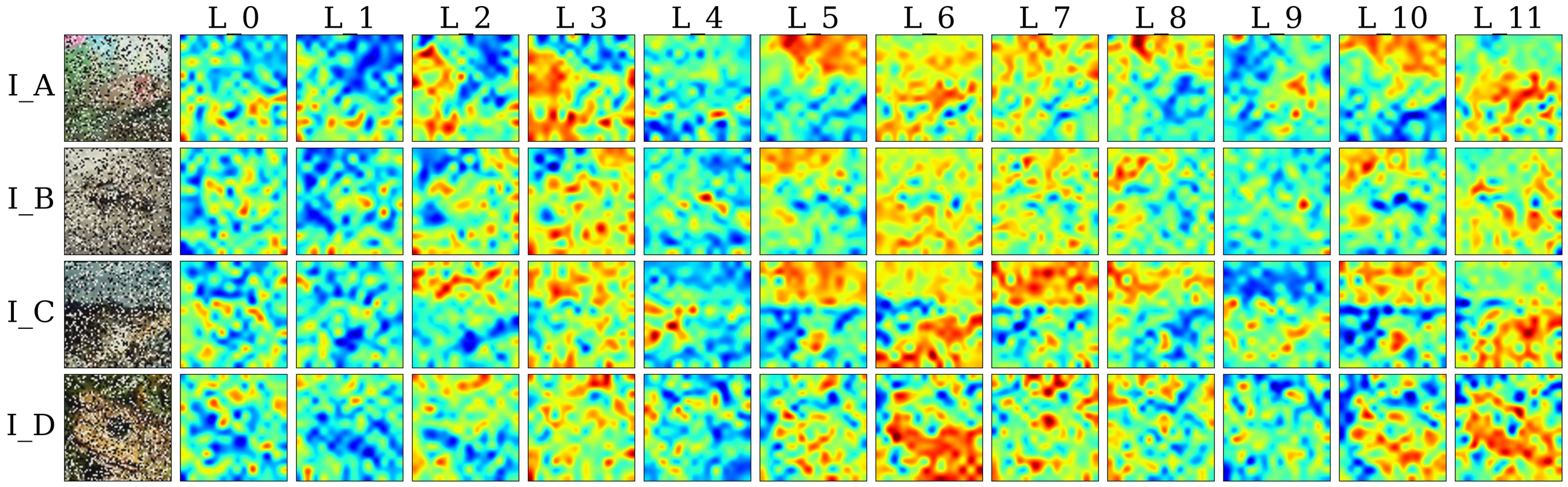}
        \caption{After training with LoRA, noisy input (30\%)}
        \label{fig:w_lora_map_after_train_noise}
    \end{subfigure}

    \caption{Layer-wise attention maps before and after familiarity training at  0\% (clean) and 30 \% noisy levels. Before familiar training, the attention map highlights accurately the critical features for object recognition {\bf a}, but becomes scattered when the input stimulus is noisy {\bf b}. Familiarity training makes the network pay attention to a large portion of the image, reflecting the encoding of more details and a broader scope of the image for remembering the entire image context, at the expense of focused attention to critical details for recognition. LoRA makes the expansion of attention more prominent ({\bf e} vs {\bf c}). On the other hand, attention maps become more robust and stable against noise than before familiar training. ({\bf c,d} vs {\bf e,f }) respectively.}
    \label{fig:attn_maps}
\end{figure*}

Figures~\ref{fig:map_before_train_clean} and \ref{fig:map_before_train_noise} show layer-wise attention maps from the pretrained CLIP encoder (prior to familiarity training) in response to clean ($0\%$ noise) and noisy ($30\%$ noise) inputs. In the clean case, early layers focus attention sharply on object regions, consistent with CLIP’s training for object-level recognition. In contrast, noise causes attention to diffuse, reflecting a breakdown in localized feature selection.

After familiarity training, this pattern shifts. As shown in Figures~\ref{fig:wo_lora_map_after_train_clean} and \ref{fig:wo_lora_map_after_train_noise}, models trained without LoRA begin to spread attention across more regions of the image, reflecting increased sensitivity to global context. Rather than focusing solely on task-relevant (object) features, the network learns to attend to broader aspects of the image context. 

This trend is even more pronounced in LoRA-trained models (Figures~\ref{fig:w_lora_map_after_train_clean} and \ref{fig:w_lora_map_after_train_noise}), which exhibit rich, spatially distributed attention patterns. This may stem from the low-rank constraints imposed by LoRA modules, which force the network to efficiently encode contextual information through limited additional capacity.

\paragraph{Quantifying Figure–Ground Sensitivity in Attention.}  
A qualitative inspection of LoRA-trained attention maps suggests improved figure–ground discrimination. To test this, we compute a layer-wise figure–ground Intersection over Union (fg-IoU) between binarized attention maps and reference segmentation masks.

Reference masks are obtained using the Grounding-SAM pipeline \cite{ren2024grounded}, which leverages Grounding DINO \cite{liu2024grounding} for object detection and SAM \cite{kirillov2023segment} for mask generation. Although our autoencoder is trained without labels, it may still exhibit emergent semantic segmentation behaviors.

\begin{figure}[htbp]
    \centering
    \begin{subfigure}[t]{0.48\linewidth}
        \centering
        \includegraphics[width=\linewidth]{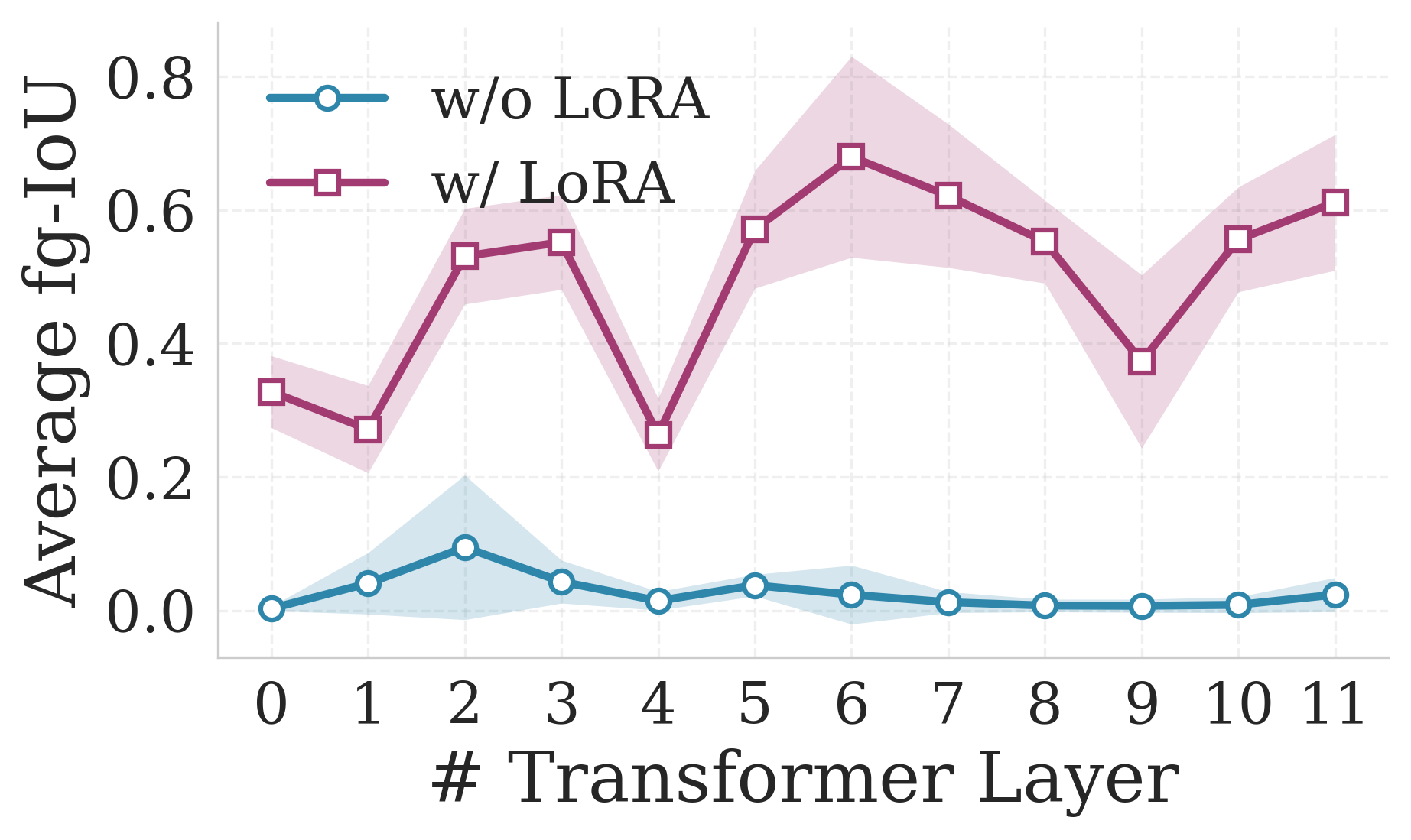}
        \caption{Layer-wise fg-IoU with reference masks}
        \label{fig:iou}
    \end{subfigure}
    \hfill
    \begin{subfigure}[t]{0.48\linewidth}
        \centering
        \includegraphics[width=\linewidth]{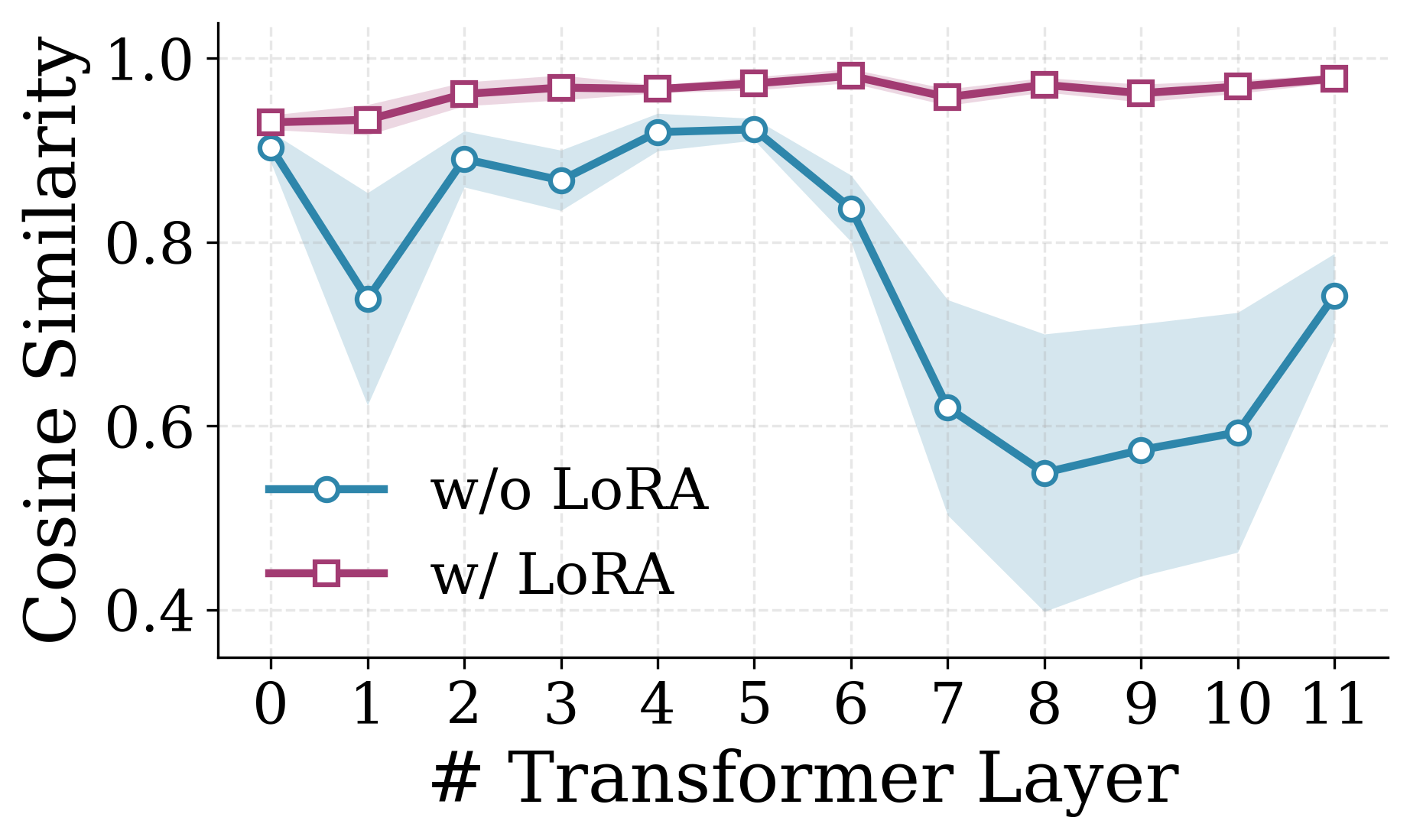}
        \caption{Cosine similarity between clean and noisy attention maps}
        \label{fig:attn_map_cos}
    \end{subfigure}
    \caption{{\bf a} Quantitative assessment of the alignment of attention map against figure-background segmentation mask. Networks with LoRA exhibit stronger figure-ground alignment. {\bf b} The alignment of different layers' attention score maps between the responses to clean images and that of the noisy images with and without LoRA.}
    \label{fig:iou_cos_combined}
\end{figure}

We define fg-IoU as:
\begin{equation}
\label{eq:iou}
\text{fg-IoU} = \max \left( \text{IoU}(\mathbf{A} > t, \mathbf{M}),\ \text{IoU}(\mathbf{A} > t, 1 - \mathbf{M}) \right)
\end{equation}

where $\mathbf{A}$ is the attention map, $\mathbf{M}$ is the binary reference mask, and $t = 0.5$ is the binarization threshold.

As shown in Figure~\ref{fig:iou}, LoRA-equipped models achieve substantially higher fg-IoU scores across layers—surpassing 0.6 in some cases—whereas non-LoRA models remain near zero. This indicates that LoRA facilitates attention structures that respect figure–ground organization.

\paragraph{Attention Stability Across Noise.}  
Another noteworthy observation is the increased similarity between attention maps for clean and noisy inputs after familiarity training, particularly in the LoRA condition. We quantify this stability using cosine similarity between attention maps across noise levels.

Figure~\ref{fig:attn_map_cos} shows that LoRA-trained models yield significantly more consistent attention across layers than both non-LoRA and untrained models. This suggests that LoRA enhances the noise-invariance of attention dynamics.

\paragraph{Interpretation.}  
Taken together, these results suggest that familiarity training encourages the network to encode detailed global context through diffuse attention patterns. This shift supports more robust representations under noise and occlusion, but also moves the network away from sparse, task-driven saliency. LoRA mitigates this risk by confining familiarity-driven plasticity to low-rank subspaces, allowing the network to potentially modulate or suppress context memories based on task demands.

Thus, LoRA not only improves representational efficiency but also offers a mechanism for compartmentalizing memory and attention—retaining the benefits of global familiarity while preserving task flexibility.

\section{Discussion}

In this paper, we propose that LoRA provides an effective framework for modeling fast weights. This approach helps to preserve the model's core generic statistical priors of natural scenes,
%maintaining its generalization ability and previously learned knowledge, 
while allowing adaptation to learning new contexts. 
% By focusing updates on a smaller set of trainable parameters, LoRA improves both data efficiency and training efficiency in learning and representing the specific familiar context. 
By focusing updates on fewer parameters, LoRA more efficiently learns and represents specific familiar contexts.

A key finding of this paper is that familiarity training brings early and intermediate encoder layers into closer alignment with the top-layer representation that encodes global contextual information. Through the autoencoder’s reconstruction loss, bottom-up features are fine-tuned to reflect top-down global structure, effectively integrating global context into early-stage representations. This process resembles analysis-by-synthesis or predictive coding: early layers increasingly mirror the global memory stored in deeper layers, supported by self-attention mechanisms that approximate flexible gating and multiplicative interactions.

However, familiarity training alone does not yield coherent figure–ground sensitive attention. Without LoRA, attention remains diffuse and lacks spatial organization. LoRA’s low-rank updates serve as an effective regularizer, enforcing spatial coherence and producing emergent grouping across regions, analogous to hypercolumn-like organization in cortex. From a neuroscience perspective, this reflects how recurrent or lateral interactions compress noise dimensions of the manifold and induce correlated activity patterns. The LoRA-enhanced model achieves stronger and more biologically consistent manifold compression than the model without LoRA, which might be a consequence of operations in the low-dimensional manifold.

In this work, we freeze only the specific modules where LoRA adapters are inserted, yielding a micro-circuit with two distinct time scales:  
(i) the frozen full-rank weights constitute \emph{slow pathways} that preserve long-term visual priors;  
(ii) the low-rank LoRA adapters act as \emph{fast, transient synaptic adjustments}, rapidly encoding task-specific information.  
The surrounding layers remain trainable, enabling global representations to reorganize around these locally stabilized modules.
This “slow-core/fast-shell’’ arrangement accords with neurophysiological evidence that cortical plasticity is both localized and circuit-specific \cite{poort2015learning, gilbert2012adult}, while still supporting co-existing slow and fast synaptic components \cite{abbott2004synaptic}.  
By combining the stability of slow pathways with the adaptability of fast weights, our partial-freezing strategy attains biologically inspired plasticity without the over-constraint imposed by fully frozen backbones.

%The capacity of a low-dimensional manifold modified via LoRA places an upper bound on the number of global memories it can encode. 
The low-dimensional manifold induced by LoRA imposes a limit on how many global memories can be encoded.
We found that increasing the number of stimuli to be represented necessitates a corresponding increase in the dimensionality of LoRA. An alternative strategy is to deploy multiple LoRA modules, each encoding a subset of global memories within its own low-dimensional manifold. 
Selection among these modules could be achieved through image-dependent routing or mixture mechanisms. The mechanisms for implementing such selective or compositional use of LoRA remain an open question for future research.

The autoencoder does not exhibit tuning-curve sharpening after familiarity training, with or without LoRA. Introducing L1 regularization to encourage sparsity 
%reduced downstream classification performance and 
attenuated global context sensitivity, indicating that simple sparsity constraints do not capture the operative principles. This divergence from neurophysiological evidence points to a deeper mechanistic distinction between recurrent neural circuit and ViT-based encoders; the manifold transform in the former is driven by selective amplification and strong normalization, which may not be aligned with the inductive bias of ViT. Clarifying how sparsity, manifold transforms, and context sensitivity interact in these models represents a tractable and important direction for future work.
%, with direct relevance to computational theories of visual coding.

In this work, the autoencoder augmented with LoRA fast weights serves as a surrogate for the hierarchical visual system equipped with recurrent circuits. The self-supervised learning process, driven by reconstruction loss, functionally approximates the rapid plasticity mechanisms thought to underlie familiarity learning in recurrent cortical connections. Notably, the manifold transformation, specifically the compression of noise dimensions, observed in both neural circuit models and the autoencoder (particularly without LoRA) appears strikingly similar. 
%This suggests that, at the level of computational theory and algorithmic function, the approximation is plausible, even if it diverges at the level of biological implementation. 
This suggests that the approximation is plausible at the computational and algorithmic levels, though it diverges at the biological implementation level.
This work thus provides insights into the computational processes and functional implications of the familiarity learning phenomena observed in the visual cortex.

\newpage
%Bibliography
\bibliographystyle{unsrt}  
\bibliography{references}

@article{hu2022lora,
  title={Lora: Low-rank adaptation of large language models.},
  author={Hu, Edward J and Shen, Yelong and Wallis, Phillip and Allen-Zhu, Zeyuan and Li, Yuanzhi and Wang, Shean and Wang, Lu and Chen, Weizhu and others},
  journal={ICLR},
  volume={1},
  number={2},
  pages={3},
  year={2022}
}

@inproceedings{radford2021learning,
  title={Learning transferable visual models from natural language supervision},
  author={Radford, Alec and Kim, Jong Wook and Hallacy, Chris and Ramesh, Aditya and Goh, Gabriel and Agarwal, Sandhini and Sastry, Girish and Askell, Amanda and Mishkin, Pamela and Clark, Jack and others},
  booktitle={International conference on machine learning},
  pages={8748--8763},
  year={2021},
  organization={PmLR}
}

@inproceedings{yang2024brain,
  title={Brain decodes deep nets},
  author={Yang, Huzheng and Gee, James and Shi, Jianbo},
  booktitle={Proceedings of the IEEE/CVF Conference on Computer Vision and Pattern Recognition},
  pages={23030--23040},
  year={2024}
}

@article{dosovitskiy2020image,
  title={An image is worth 16x16 words: Transformers for image recognition at scale},
  author={Dosovitskiy, Alexey and Beyer, Lucas and Kolesnikov, Alexander and Weissenborn, Dirk and Zhai, Xiaohua and Unterthiner, Thomas and Dehghani, Mostafa and Minderer, Matthias and Heigold, Georg and Gelly, Sylvain and others},
  journal={arXiv preprint arXiv:2010.11929},
  year={2020}
}

@article{poort2015learning,
  title={Learning enhances sensory and multiple non-sensory representations in primary visual cortex},
  author={Poort, Jasper and Khan, Adil G and Pachitariu, Marius and Nemri, Abdellatif and Orsolic, Ivana and Krupic, Julija and Bauza, Marius and Sahani, Maneesh and Keller, Georg B and Mrsic-Flogel, Thomas D and others},
  journal={Neuron},
  volume={86},
  number={6},
  pages={1478--1490},
  year={2015},
  publisher={Elsevier}
}

@article{gilbert2012adult,
  title={Adult visual cortical plasticity},
  author={Gilbert, Charles D and Li, Wu},
  journal={Neuron},
  volume={75},
  number={2},
  pages={250--264},
  year={2012},
  publisher={Elsevier}
}

@article{abbott2004synaptic,
  title={Synaptic computation},
  author={Abbott, Larry F and Regehr, Wade G},
  journal={Nature},
  volume={431},
  number={7010},
  pages={796--803},
  year={2004},
  publisher={Nature Publishing Group UK London}
}

@inproceedings{liu2024grounding,
  title={Grounding dino: Marrying dino with grounded pre-training for open-set object detection},
  author={Liu, Shilong and Zeng, Zhaoyang and Ren, Tianhe and Li, Feng and Zhang, Hao and Yang, Jie and Jiang, Qing and Li, Chunyuan and Yang, Jianwei and Su, Hang and others},
  booktitle={European conference on computer vision},
  pages={38--55},
  year={2024},
  organization={Springer}
}

@inproceedings{kirillov2023segment,
  title={Segment anything},
  author={Kirillov, Alexander and Mintun, Eric and Ravi, Nikhila and Mao, Hanzi and Rolland, Chloe and Gustafson, Laura and Xiao, Tete and Whitehead, Spencer and Berg, Alexander C and Lo, Wan-Yen and others},
  booktitle={Proceedings of the IEEE/CVF international conference on computer vision},
  pages={4015--4026},
  year={2023}
}

@article{ren2024grounded,
  title={Grounded sam: Assembling open-world models for diverse visual tasks},
  author={Ren, Tianhe and Liu, Shilong and Zeng, Ailing and Lin, Jing and Li, Kunchang and Cao, He and Chen, Jiayu and Huang, Xinyu and Chen, Yukang and Yan, Feng and others},
  journal={arXiv preprint arXiv:2401.14159},
  year={2024}
}

@article{meyer2014image,
  title={Image familiarization sharpens response dynamics of neurons in inferotemporal cortex},
  author={Meyer, Travis and Walker, Christopher and Cho, Raymond Y and Olson, Carl R},
  journal={Nature neuroscience},
  volume={17},
  number={10},
  pages={1388--1394},
  year={2014},
  publisher={Nature Publishing Group US New York}
}

@article{fahy1993neuronal,
  title={Neuronal activity related to visual recognition memory: long-term memory and the encoding of recency and familiarity information in the primate anterior and medial inferior temporal and rhinal cortex},
  author={Fahy, FL and Riches, IP and Brown, MW},
  journal={Experimental Brain Research},
  volume={96},
  pages={457--472},
  year={1993},
  publisher={Springer}
}

@article{xiang1998differential,
  title={Differential neuronal encoding of novelty, familiarity and recency in regions of the anterior temporal lobe},
  author={Xiang, J-Z and Brown, MW},
  journal={Neuropharmacology},
  volume={37},
  number={4-5},
  pages={657--676},
  year={1998},
  publisher={Elsevier}
}

@article{mruczek2007context,
  title={Context familiarity enhances target processing by inferior temporal cortex neurons},
  author={Mruczek, Ryan EB and Sheinberg, David L},
  journal={Journal of Neuroscience},
  volume={27},
  number={32},
  pages={8533--8545},
  year={2007},
  publisher={Soc Neuroscience}
}

@article{sobotka1993investigation,
  title={Investigation of long term recognition and association memory in unit responses from inferotemporal cortex},
  author={Sobotka, Stanislaw and Ringo, James L},
  journal={Experimental Brain Research},
  volume={96},
  pages={28--38},
  year={1993},
  publisher={Springer}
}

@article{freedman2006experience,
  title={Experience-dependent sharpening of visual shape selectivity in inferior temporal cortex},
  author={Freedman, David J and Riesenhuber, Maximilian and Poggio, Tomaso and Miller, Earl K},
  journal={Cerebral Cortex},
  volume={16},
  number={11},
  pages={1631--1644},
  year={2006},
  publisher={Oxford University Press}
}

@article{woloszyn2012effects,
  title={Effects of long-term visual experience on responses of distinct classes of single units in inferior temporal cortex},
  author={Woloszyn, Luke and Sheinberg, David L},
  journal={Neuron},
  volume={74},
  number={1},
  pages={193--205},
  year={2012},
  publisher={Elsevier}
}

@article{huang2018neural,
  title={Neural correlate of visual familiarity in macaque area V2},
  author={Huang, Ge and Ramachandran, Suchitra and Lee, Tai Sing and Olson, Carl R},
  journal={Journal of Neuroscience},
  volume={38},
  number={42},
  pages={8967--8975},
  year={2018},
  publisher={Soc Neuroscience}
}

@article{wang2025manifold,
  title={Manifold transform by recurrent cortical circuit enhances robust encoding of familiar stimuli},
  author={Wang, Weifan and Niu, Xueyan and Liang, Liyuan and Lee, Tai-Sing},
  journal={PLOS Computational Biology},
  volume={21},
  number={10},
  pages={e1013587},
  year={2025},
  publisher={Public Library of Science San Francisco, CA USA}
}

@inproceedings{hinton1987using,
  title={Using fast weights to deblur old memories},
  author={Hinton, Geoffrey E and Plaut, David C},
  booktitle={Proceedings of the ninth annual conference of the Cognitive Science Society},
  pages={177--186},
  year={1987}
}

@article{ba2016using,
  title={Using fast weights to attend to the recent past},
  author={Ba, Jimmy and Hinton, Geoffrey E and Mnih, Volodymyr and Leibo, Joel Z and Ionescu, Catalin},
  journal={Advances in neural information processing systems},
  volume={29},
  year={2016}
}

@article{hopfield1982neural,
  title={Neural networks and physical systems with emergent collective computational abilities.},
  author={Hopfield, John J},
  journal={Proceedings of the national academy of sciences},
  volume={79},
  number={8},
  pages={2554--2558},
  year={1982}
}

@article{ramsauer2020hopfield,
  title={Hopfield networks is all you need},
  author={Ramsauer, Hubert and Sch{\"a}fl, Bernhard and Lehner, Johannes and Seidl, Philipp and Widrich, Michael and Adler, Thomas and Gruber, Lukas and Holzleitner, Markus and Pavlovi{\'c}, Milena and Sandve, Geir Kjetil and others},
  journal={arXiv preprint arXiv:2008.02217},
  year={2020}
}

@article{rao1999predictive,
  title={Predictive coding in the visual cortex: a functional interpretation of some extra-classical receptive-field effects},
  author={Rao, Rajesh PN and Ballard, Dana H},
  journal={Nature neuroscience},
  volume={2},
  number={1},
  pages={79--87},
  year={1999},
  publisher={Nature Publishing Group}
}

@article{gilbert2013top,
  title={Top-down influences on visual processing},
  author={Gilbert, Charles D and Li, Wu},
  journal={Nature reviews neuroscience},
  volume={14},
  number={5},
  pages={350--363},
  year={2013},
  publisher={Nature Publishing Group UK London}
}

@article{coen2015flexible,
  title={Flexible gating of contextual influences in natural vision},
  author={Coen-Cagli, Ruben and Kohn, Adam and Schwartz, Odelia},
  journal={Nature neuroscience},
  volume={18},
  number={11},
  pages={1648--1655},
  year={2015},
  publisher={Nature Publishing Group US New York}
}

@article{vaswani2017attention,
  title={Attention is all you need},
  author={Vaswani, Ashish and Shazeer, Noam and Parmar, Niki and Uszkoreit, Jakob and Jones, Llion and Gomez, Aidan N and Kaiser, {\L}ukasz and Polosukhin, Illia},
  journal={Advances in neural information processing systems},
  volume={30},
  year={2017}
}

@article{le2015tiny,
  title={Tiny imagenet visual recognition challenge},
  author={Le, Yann and Yang, Xuan},
  journal={CS 231N},
  volume={7},
  number={7},
  pages={3},
  year={2015}
}

@article{lee2003hierarchical,
  title={Hierarchical Bayesian inference in the visual cortex},
  author={Lee, Tai Sing and Mumford, David},
  journal={Journal of the Optical Society of America A},
  volume={20},
  number={7},
  pages={1434--1448},
  year={2003},
  publisher={Optical Society of America}
}

@article{angelucci2017circuits,
  title={Circuits and mechanisms for surround modulation in visual cortex},
  author={Angelucci, Alessandra and Bijanzadeh, Maryam and Nurminen, Lauri and Federer, Frederick and Merlin, Sam and Bressloff, Paul C},
  journal={Annual review of neuroscience},
  volume={40},
  number={1},
  pages={425--451},
  year={2017},
  publisher={Annual Reviews}
}

@article{geisler2008visual,
  title={Visual perception and the statistical properties of natural scenes},
  author={Geisler, Wilson S},
  journal={Annu. Rev. Psychol.},
  volume={59},
  number={1},
  pages={167--192},
  year={2008},
  publisher={Annual Reviews}
}

@article{yan2018bottom,
  title={Bottom-up saliency and top-down learning in the primary visual cortex of monkeys},
  author={Yan, Yin and Zhaoping, Li and Li, Wu},
  journal={Proceedings of the National Academy of Sciences},
  volume={115},
  number={41},
  pages={10499--10504},
  year={2018},
  publisher={National Academy of Sciences}
}

@article{mcclelland1995there,
  title={Why there are complementary learning systems in the hippocampus and neocortex: insights from the successes and failures of connectionist models of learning and memory.},
  author={McClelland, James L and McNaughton, Bruce L and O'Reilly, Randall C},
  journal={Psychological review},
  volume={102},
  number={3},
  pages={419},
  year={1995},
  publisher={American Psychological Association}
}

@article{tang2018large,
  title={Large-scale two-photon imaging revealed super-sparse population codes in the V1 superficial layer of awake monkeys},
  author={Tang, Shiming and Zhang, Yimeng and Li, Zhihao and Li, Ming and Liu, Fang and Jiang, Hongfei and Lee, Tai Sing},
  journal={Elife},
  volume={7},
  pages={e33370},
  year={2018},
  publisher={eLife Sciences Publications, Ltd}
}

\end{document}